\documentclass[conference]{IEEEtran}
\IEEEoverridecommandlockouts
\makeatletter
\def\ps@IEEEtitlepagestyle{%
  \def\@oddfoot{\mycopyrightnotice}%
  \def\@oddhead{\hbox{}\@IEEEheaderstyle\leftmark\hfil\thepage}\relax
  \def\@evenhead{\@IEEEheaderstyle\thepage\hfil\leftmark\hbox{}}\relax
  \def\@evenfoot{}%
}
\def\mycopyrightnotice{%
  \begin{minipage}{\textwidth}
  \centering \scriptsize
  Copyright~\copyright~2024 IEEE. Personal use of this material is permitted. Permission from IEEE must be obtained for all other uses, in any current or future media, including\\reprinting/republishing this material for advertising or promotional purposes, creating new collective works, for resale or redistribution to servers or lists, or reuse of any copyrighted component of this work in other works.
  \end{minipage}
}
\makeatother
\usepackage[noadjust]{cite}
\usepackage{multicol}
\usepackage{multirow}
\usepackage{booktabs} 
\usepackage{amsmath,amssymb,amsfonts,amsthm}
\usepackage{textcomp}
\usepackage{xcolor}
\usepackage{graphicx}
\usepackage[caption=false,font=footnotesize]{subfi
g}
\def\BibTeX{{\rm B\kern-.05em{\sc i\kern-.025em b}\kern-.08em
    T\kern-.1667em\lower.7ex\hbox{E}\kern-.125emX}}

\begin{document}

\title{CoRAST: Towards Foundation Model-Powered Correlated Data Analysis in  Resource-Constrained CPS and IoT
\thanks{This work was funded by the  NSF grant CNS-2106891 and the Wei Shen and Xuehong Zhang Presidential Fellowship at Carnegie Mellon University.}
}
\newcommand{\Xname}{CoRAST }
\newcommand{\ctxt}[1]{\raisebox{.5pt}{\textcircled{\raisebox{-.9pt} {#1}}}}

\author{\IEEEauthorblockN{Yi Hu\IEEEauthorrefmark{1}, Jinhang Zuo\IEEEauthorrefmark{2}, Alanis Zhao\IEEEauthorrefmark{1}, Bob Iannucci\IEEEauthorrefmark{3} and Carlee Joe-Wong\IEEEauthorrefmark{1}} 

\IEEEauthorblockA{\IEEEauthorrefmark{1}Carnegie Mellon University, \IEEEauthorrefmark{2}Caltech \& UMass Amherst, \IEEEauthorrefmark{3}Google }

\IEEEauthorblockA{\{yihu,aazhao,cjoewong\}@andrew.cmu.edu, jhzuo@cs.umass.edu, biannucci@google.com}
}

\maketitle

\begin{abstract}
Foundation models (FMs) emerge as a promising solution to harness distributed and diverse environmental data by leveraging prior knowledge to understand the complicated temporal and spatial correlations within heterogeneous datasets.  Unlike distributed learning frameworks such as federated learning, which often struggle with multimodal data, FMs can transform diverse inputs into embeddings. This process facilitates the integration of information from various modalities and the application of prior learning to new domains. However, deploying FMs in resource-constrained edge systems poses significant challenges.  To this end, we introduce \textbf{CoRAST}, a novel learning framework that utilizes  FMs  for enhanced analysis of distributed, correlated heterogeneous data. Utilizing a server-based FM, CoRAST exploits existing environment information to extract temporal, and cross-feature correlations  among sensor data. This enables CoRAST to offer context-aware insights for localized client tasks through FM-powered global representation learning. Our evaluation on real-world weather dataset demonstrates CoRAST's ability to exploit correlated heterogeneous data through environmental representation learning  to reduce the forecast errors by up to $50.3\%$ compared to the baselines.
\end{abstract}

\begin{IEEEkeywords}
Cyber-physical systems, foundation models, heterogeneous data analysis, Internet of Things (IoT), time series.
\end{IEEEkeywords}

\section{Introduction}
\begin{figure}
    \centering
    \includegraphics[width=.75\linewidth]{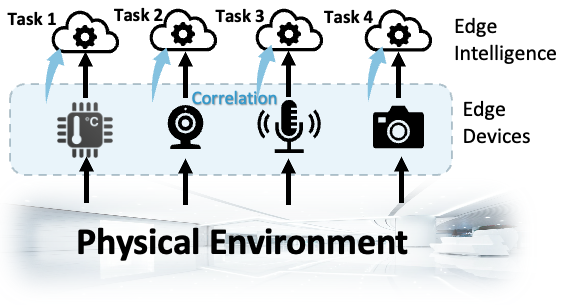}
    \caption{Correlated Data in CPS and IoT}
    \label{fig:demo}
\end{figure}

The data generated by edge devices in Cyber-Physical Systems (CPS) and the Internet of Things (IoT) is inherently rich in spatial and temporal correlations, offering significant opportunities for enhancing edge intelligence. Those systems' distributed computing and sensing capabilities enable the monitoring of a physical environment through different modalities (e.g., audio/video signals,  or environmental metrics like temperature and air pressure). Effectively utilizing this rich and interrelated data pool, collected from a diverse set of edge sensors, can substantially benefit a wide range of complex downstream tasks that require detailed data interpretation (Figure \ref{fig:demo}). For example, augmenting audio data with visual cues (e.g., lip-reading) from a camera focusing on a speaker  can significantly improve the accuracy of speech recognition~\cite{ryumin2023audio}.

However, it is challenging to leverage this rich, distributed data effectively. Extensive research has been done on multivariate time series for forecasting or classification, and the major approach involves treating each variable as an independent univariate time series, addressing temporal and inter-variable correlations separately~\cite{zhao2020mtad-gat,darban2022deep,chen2021gta}. There is a lack of a unified strategy for integrating heterogeneous data types, such as video streams and time series data. 
Moreover, the prevalent practice of centralizing data processing can overlook essential spatial details. For instance, when one sensor captures 1D acceleration data at a particular location, while another records 3D gyroscope data elsewhere, the central system may fail to fully leverage the spatial context of these data points. This limitation underscores the need for strategies that can effectively integrate and interpret heterogeneous data sources while preserving and utilizing their spatial characteristics to enhance analysis and decision-making.

Foundation models (FMs) emerge a promising solution  to complex data analysis and information fusion. Research in this domain is rapidly gaining popularity, as evidenced  by recent surveys~\cite{min2023recent,naveed2023comprehensive,jin2023large,wu2023multimodal}. Employing transformers, these models adeptly convert diverse inputs into embeddings, enabling the \textit{integration of various data types and facilitating the application of prior learning or training to new domains}.  Large Language Models (LLMs) like GPT and BERT extend well beyond their initial applications in text and image processing to encompass temporal and multimodal data analysis. Research by Zhou et al.~\cite{Zhou2023OneFA} exemplifies how pre-trained LLMs can match or even outperform the state-of-the-art methods in time series analysis.  Advanced multimodal models ~\cite{lyu2023macaw,radford2021clip} further integrate multimodal data -- from images and text to audio and video -- for a diverse range of downstream tasks. 


Nonetheless, integrating FMs into CPS and IoT applications poses major {\textbf{challenges}}, primarily due to the extensive resources required to run these large models, often beyond the capabilities of edge devices. 
Many CPS and IoT applications require edge devices themselves to make decisions or predictions. For instance, in smart agriculture, soil moisture sensors may need to determine optimal irrigation levels based on collected data. These devices, e.g., low-power sensors, may not have the communication capabilities needed to continually send collected data to a central location and receive back a decision. Moreover, sending some data, e.g., audiovisual data, to a central location for processing may introduce privacy issues.
We aim to answer the question: \textit{how can we leverage the knowledge from a foundation model, previously trained on historical data, for distributed systems with resource-constrained edge devices?} 
Research  {challenges} include:

\textbf{Resource Constraints on Edge Devices.} 
Edge devices or sensors within CPS or IoT environments often face severe memory and computational limitations, making it difficult to host FMs directly. For example, the base BERT model~\cite{devlin2019bert}  has 108 million  parameters. Training even moderately sized FMs typically requires several  gigabytes of memory, presenting a challenge for updating these models on edge devices. This is true even when considering distilled or compressed versions of FMs designed to reduce  size and resource requirements~\cite{sanh2019distilbert}. High-end edge computing devices, such as NVIDIA Jetsons, might struggle with the storage and memory demands required for training large models, let alone less capable IoT devices (e.g., microcontrollers) that are typically equipped with only a few megabytes of SRAM. Thus, involving FMs in training on locally collected environmental data may require significant involvement of more well-resourced servers that can host FMs.

\textbf{Decentralized and Correlated Local Tasks.} 
CPS and IoT applications often need distributed decision-making on the edge due to privacy concerns, real-time responsiveness, and unreliable network connections.  This distributed intelligence calls for the development of local models tailored to the specific edge data models and distributions. Traditional federated learning (FL), even with personalization, often falls short when it comes to learning with correlated and heterogeneous local data. For example, adjacent edge devices with different sensor arrays may collect correlated yet distinct types of data, calling for unique local models for tasks like anomaly detection. This level of data specificity and the need for model customization go beyond what personalized federated learning currently offers. In light of the fact that many edge devices cannot directly host FMs, we must design new methods to integrate information from FMs hosted at a central server into distributed decision-making on such correlated data, without leading to excessive communication overhead. 

\paragraph*{{Contributions}} To address these research challenges, we introduce \textbf{CoRAST}, a novel general learning framework designed to utilize FMs for \textbf{A}nalyzing \textbf{S}patially and \textbf{T}emporally \textbf{CoR}related heterogeneous data  within resource-constrained edge computing systems. Diverging from traditional approaches based off improving a single application objective, CoRAST adopts a holistic strategy, emphasizing the interconnectedness of data from a shared physical environment, thereby improving the learning for various downstream tasks. Leveraging FMs to  understand the correlation among various sensor data and tasks with prior environmental knowledge, CoRAST provides  context-aware insights for client-specific tasks. We summarize our contributions:
\begin{itemize}
    \item We introduce CoRAST, the first FM-based learning framework for analyzing correlated heterogeneous data in CPS and IoT. The framework leverage a server-based FM to synthesize global insights and client-specific models that apply these insights to local tasks.
    \item Our study offers insights into designing FM-based architectures for managing correlated heterogeneous data.
    \item CoRAST improves distributed learning on a real-world weather dataset, reducing forecasting errors  with its FM-based global learning approach. 
\end{itemize}

{We review related work in Section \ref{sec:related} and present CoRAST in Section \ref{sec:system}. The proposed framework is evaluated in Section \ref{sec:evaluation}. The paper concludes in Section \ref{sec:conclusion}. }




\section{Related Work}\label{sec:related}

\paragraph*{FM for Correlated and Multimodal Data Analysis} Research on FM for correlated data analysis is rapidly advancing with models such as GPT4TS~\cite{Zhou2023OneFA}, LLM4TS~\cite{chang2024llm4ts}, and AnomalyTrans~\cite{wen2023transformers}, which utilize pre-trained models and transformers for tasks like time series analysis, forecasting, and anomaly detection. These models enhance their performance through targeted fine-tuning. GTA~\cite{chen2021gta} employing graph structures for uncovering hidden data associations with novel attention mechanisms. Penetrative AI~\cite{xu2024penetrative} exploits textualized expert knowledge to guide the reasoning of the physical world using existing language models, but it has limitations in detailed data analysis and comprehensive environment learning. Additionally, the adoption of multimodal large models, including CLIP~\cite{radford2021clip} and Macaw-LLM~\cite{lyu2023macaw}, demonstrates FMs' potential to unify data modalities such as in text-image translation by converting diverse inputs into embeddings. 

\paragraph*{Advancements in Representation Learning} TS2Vec\cite{yue2022ts2vec} introduces a universal contrastive learning approach  {to time series representations} that effectively differentiates between positive and negative samples at various time scales with  contextual information. In federated learning, CreamFL~\cite{yu2023multimodal} pushes the boundaries of federated multimodal learning by sharing learned modality-specific representations.  FedGKT~\cite{he2020group} explores a novel strategy for  knowledge transfer using local representation and global logits. In contrast, CoRAST explicitly incorporates FMs to learn representations capturing correlations between distributed and multi-modal data.
\begin{figure}
    \centering
    \includegraphics[width=\linewidth]{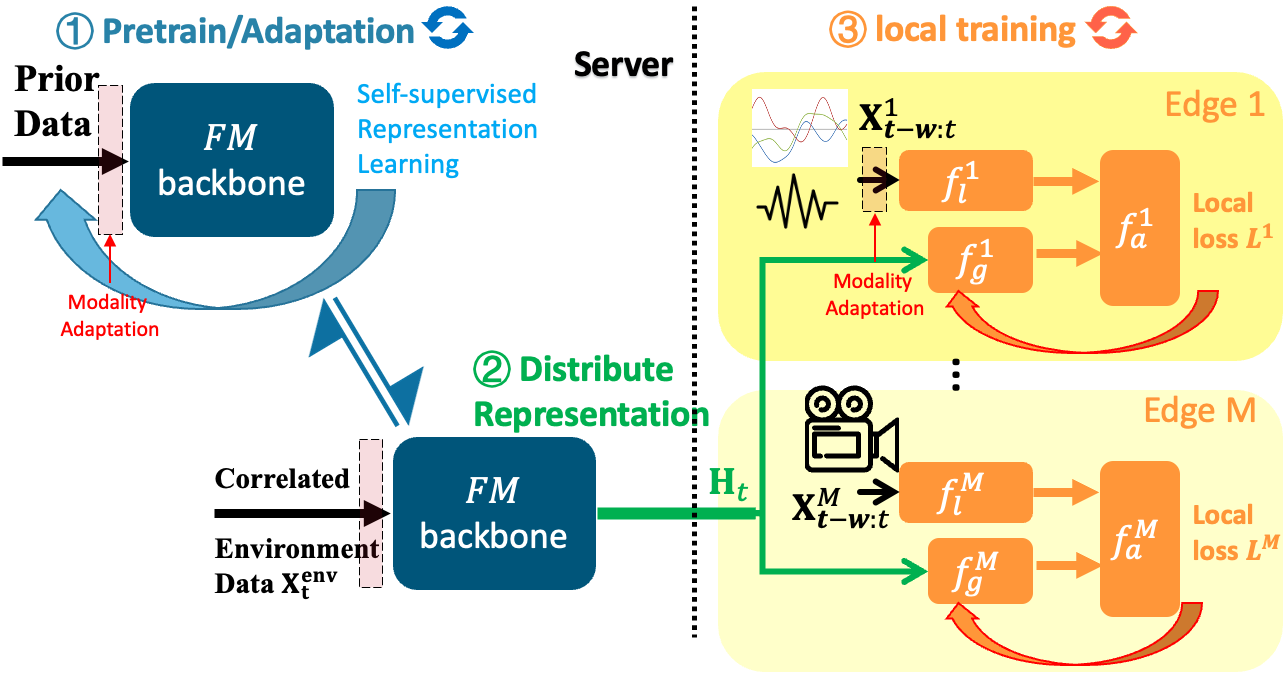}
    \caption{CoRAST framework.  \ctxt{1} Global Representation Learning -- A server-based FM is pre-trained and fine-tuned on historical environment data for global representation learning in a self-supervised manner. \ctxt{2} Representation Distribution -- Using  the environment data that correlate with  client local data,  the FM produces and distributes contextual representations to edge clients aiding downstream local tasks.  \ctxt{3} Local Learning with Global Context -- Clients integrate global contexts with their local data for independent local learning, leveraging broad environmental insights with local datasets.   }
    \label{fig:framework}
\end{figure}

\section{C{o}RAST}\label{sec:system} CoRAST leverages a server-based FM backbone to exploit both existing data and prior knowledge of the physical environment. It aims to learn useful representations of the environment in a self-supervised manner. These representations learned by CoRAST serve to provide valuable context for CPS and IoT applications operating within the same {physical} setting. On the client side, CoRAST integrates this environmental understanding, as learned from the server FM, with local data to address specific edge computing tasks.

Figure \ref{fig:framework} provides an overview of the framework. We consider a distributed system of $M$ edge clients indexed by $m$, each equipped with various sensors for distinct local tasks. Each client $m$ maintains a dataset $\mathbf{X}_t^m$ and aims to optimize a local model to minimize its unique loss function $L_t^m${, which may reflect its specific local task (e.g., predicting various environmental phenomena)}. 

An edge server with the sufficient resources  to run an FM, holds a dataset $\mathbf{X}_t^{\text{env}}$ correlating with the clients' local data. The FM,  trained in a self-supervised or unsupervised way, generates global representations $\mathbf{H}_t$ of the environment to be distributed to the edge clients. This paper primarily focuses on the genral framework's design, leaving the detailed FM architecture and its adaptation to specific systems, applications, and data types for future exploration. {Our framework is agnostic to the specific architecture of this FM.}

Upon receiving the global representation $\mathbf{H}_t$ from the server, edge clients merge this information with their local data to refine their models.  To illustrate the idea, consider a scenario where the environment of interest is a single room monitored by various sensors with distinct functions (e.g., an infrared camera that detects human at the front door, a microphone designed to pick up specific keywords, and temperature sensors programmed to regulate the room's air conditioning). The server might have access to the video streams from a surveillance camera or an aggregation of sensor data to train the FM. 
The server can pre-train the FM with all the existing environmental data and generate useful environment representations $\mathbf{H}_t$  to each client. This representation embeds vital extraneous global information not directly  observable by local sensors (e.g., the presence of a person in the room), but crucial for enhancing local task performance.

By training the local model on both the specific local data and the enriched global representation, the model is equipped with a broader understanding of the environment. This comprehensive input significantly elevates the model's potential to excel in its designated task.

\subsection{Server-Based Representation Learning} The goal of the server-based representation learning is to encode the intricate relationships and correlations among multimodal data gathered from a specific environment.  Correlation describes how  data relate to each other.  If two things are independent, they are not correlated.  In the realm of information theory, discerning the correlational dynamics within data can effectively diminish system entropy -- a metric quantifying uncertainty. Should two variables $X$ and $Y$ exhibit correlation, their joint entropy $H(X,Y)$ would be less than the sum of their separate entropies $H(X)+H(Y)$~\cite{shannon1948mathematical}.  This implies that correlated data inherently reduce system uncertainty, potentially enhancing model performance when such correlations are adeptly leveraged.

Modeling correlations has been challenging. Traditional methods use meta-data, such as spatial and temporal information, or entropy analysis to discern correlations~\cite{nguyen2018entropy}.   However, they fall short in complex scenarios where the correlation changes over time, or when dealing with high-dimensional, multimodal data. In recent years, the shift towards deep learning approaches, particularly the use of large foundation models, enable autonomously derive insights from extensive datasets and learn valuable embeddings for multimodal data.

In CoRAST, we employ a  FM backbone for representation learning. This learning process is conducted in a self-supervised manner, aiming to extract correlation-centric, high-level  concepts that help improve performance across a range of downstream tasks. One notable strategy  is contrastive representation learning, which improves the model by differentiating among varied data samples~\cite{le2020contrastive}.  TS2Vec\cite{yue2022ts2vec}, for example, applies this method to time series data, capturing contextual information across various temporal scales effectively. 

Yet, correlations {in CoRAST's setting} appear as temporal (time-based changes), spatial (spatial distribution), and cross-modal (across different data types).  Addressing these correlations effectively  demands careful architectural  design. We briefly discuss two advanced techniques that enhance the ability to interpret complex system inputs and diverse data types {and thus handle more complex correlations}. 

\paragraph{Graph Neural Networks (GNNs)} GNNs stand out for their ability to  model relational data. In GNNs, node representations are iteratively updated through a series of message passing with neighbours~\cite{kipf2016semi}. {The graph structures capture complex relationships within data, such as inter-temporal dependencies--where nodes represent distinct time points, or inter-variable connections--where nodes represent different variables~\cite{zhao2020mtad-gat}. Edges thus represent temporal and inter-variable correlations. } For datasets of heterogeneous data types, heterogeneous GNNs is a viable solution that supports processing input features of varying sizes. 

\paragraph{Attention Mechanism}
Given a task, an attention score is computed, indicating the relevance between pairs of  inputs and generates embeddings based on that score~\cite{vaswani2017attention}. In our context, this process enables models to interpret the interconnections between different data features and  their temporal connections over time.  Transformers heavily rely on attention to encode  input sequences. When integrated with GNNs, it leads to the  Graph Attention Networks (GAT)~\cite{velickovic2017graph} that can discern  the complex interrelationships present within graph-structured data, enabling the extraction of meaningful representations out of spatially and temporally correlated data.  

\subsection{Client Local Training} Each client $m$'s  local model contains three parts: a local data processing module $f_l^m$, a global information extraction module $f_g^m$ that deciphers the encoded global representation $\mathbf{H}$
, and an affiliate function $f_a^m$ that integrates outputs from both modules to produce the final outputs $\mathbf{Y}^m$. Mathematically,   $\mathbf{Y}^m = f_a^m\left(f_l^m(\mathbf{X}^m), f_g^m(\mathbf{H})\right)$.

The local loss $L^m$ can be determined by the specific local tasks of each client, which might include, for instance, the use of cross-entropy loss $L_{CE}(\mathbf{Y}^m,y^m)$ for classification tasks, where $y^m$ represents the target label; or the prediction error $L_{MSE}(\mathbf{Y}^m, \mathbf{X}_f^m)$ on the discrepancy between the predicted and actual future values of the local measurements $\mathbf{X}_f^m$.  

In addition to flexibly supporting a wide range of local tasks, the client local model also allows multimodal data inputs that may exhibit temporal or cross-modal correlations. The data processing unit can be customized to capture time-dependent patterns or inter-feature correlations. For instance, temporal convolutional layers~\cite{bai2018empirical} can be used to grasp the temporal dynamics of the data in CPS and IoT systems, and attention mechanisms with graph structures can be adopted to learn the interconnections between two types of data. 

\subsection{Continual Learning}  
The CoRAST framework is designed to foster continual learning, allowing for the adaptation and incorporation of new data over time.  At its core, the framework maintains a separation between the process of learning  environmental representations and the specific client local tasks. This division   allows environmental representation learning to universally augment both present and future CPS and IoT tasks within a given environment.  Consequently, both the server model and the local client models can efficiently integrate updates or new data without the need for a complete system-wide retraining.

The framework allows different update frequencies for the server model and the client models, a practical feature  due to the usual delay in client data becoming available to the server.  One learning scheme is to have different update intervals, $T_s$ for the server model and $T_c$ for the client models, with $T_s$ being significantly longer than $T_c$. We don't focus on the specifics of how the model adjusts to new data. In the intervals between  server model updates, clients can manage incoming new data using the representations generated from previously learned server model. Clients are required to seek updated global representations from the server only when the server model is updated, an event that occurs much less frequently.  

\subsection{Resource Usage} CoRAST optimizes resource usage by enabling distributed clients to leverage the knowledge of a pre-trained FM without local execution of the large model. Instead, clients only need to run moderately-sized local models, ensuring a lightweight computational footprint. Communication overhead predominantly involves one-directional data flow from the server to the clients, mainly for the purpose of transmitting representations.  Currently, we consider a single global representation for all the clients, with the representation's size  dependent on the volume and variety of data available to the server.

For training with new data spanning a length of $T$, the global representation is  a  $d\times T$ matrix for a $d$-dimensional representation model. This representation is distributed to all clients at intervals of $T_s$, the update interval of the server model. During  inference, a $d$-dimensional vector representation is broadcast to all clients at each inference point.   

 {A future direction would involve developing customized representations tailored to individual client needs, thereby encoding only the pertinent partial information. This would further optimize computational and communication efficiency, for instance, by utilizing smaller representations for simple tasks like automatically adjusting air conditioning based on temperature. More discussion in Section \ref{sec:conclusion}. }

\begin{figure*}
    \centering
    \includegraphics[width=.9\linewidth]{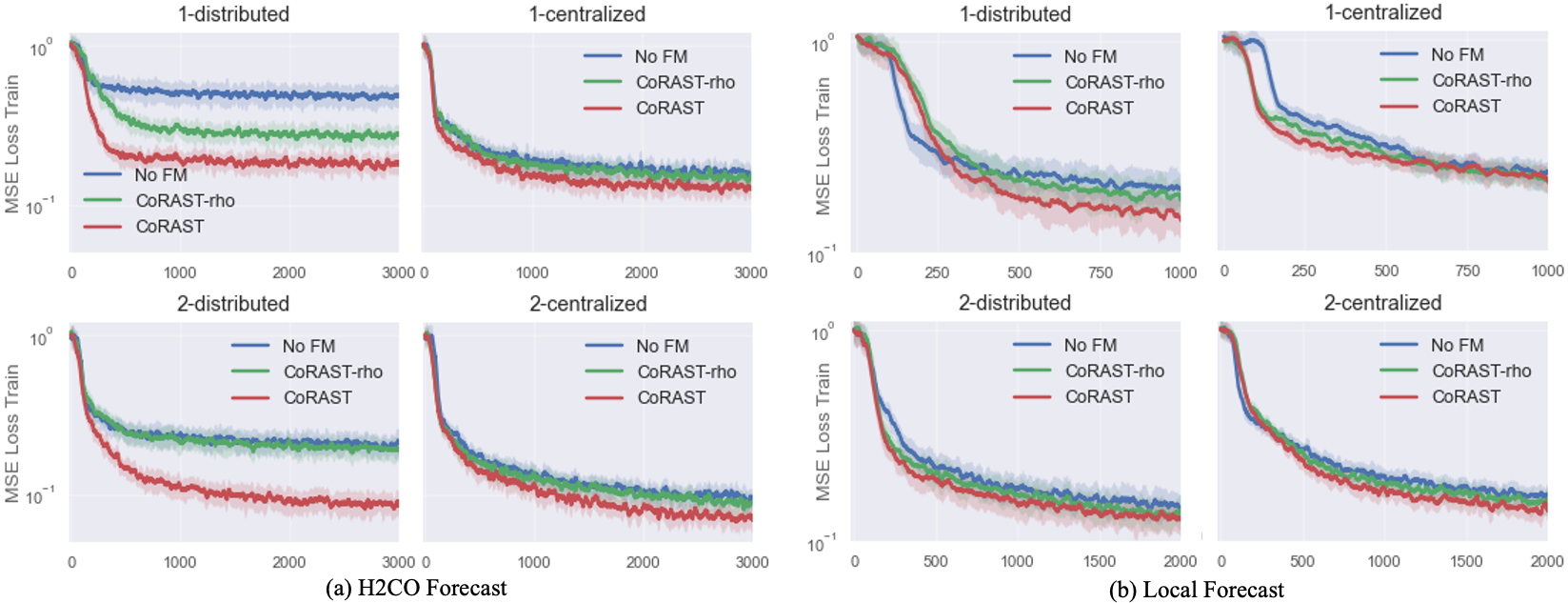}
    \caption{MSE Train Loss with the standard deviation represented by a shaded area around the loss curves. CoRAST has the lowest training loss in all settings when clients share the same local training task or have distinct local training tasks.}
    \label{fig:results}
\end{figure*}

\begin{table}
    \centering
    \caption{Experiment Settings}
    \label{tab:setting}
    \begin{minipage}{\columnwidth}
    \begin{center}
    \begin{tabular}{c c c}\toprule
       Settings  &  {Client Data} & Server Data\\\midrule
        1-centralized & [Tdew, rh, sh] & [Tdew, rh, sh]  \\
        1-distributed & [Tdew], [rh], [sh] &[Tdew, rh, sh] \\\hline
        2-centralized & [Tdew, Tpot, rh, p, sh]  &[Tdew, Tpot, rh, p, sh]\\
        2-distributed & [Tdew, Tpot], [rh, p], [sh] &[Tdew, Tpot, rh, p, sh]\\\hline
    \end{tabular}
    \end{center}
    \end{minipage}
\end{table}

\begin{table}
    \centering
    \caption{Model Setup}
    \label{tab:setup}
    \begin{minipage}{\columnwidth}
    \begin{center}
    \begin{tabular}{c c c }\toprule
       {Model} & & \# of parameters  \\\midrule
       \multirow{4}{*}{Client (TCN)} & 1-centralized &  13984  \\
        &2-centralized &  15214   \\
&1-distributed &  12970, 12970, 12970   \\
&2-distributed &  13450, 13450, 12970   \\\midrule
{Server (TS2Vec)} & &337152\\\midrule
{Representation} & & 256\\\bottomrule[\heavyrulewidth]

    \end{tabular}
    \end{center}
    \end{minipage}
\end{table}

\section{Experiments}\label{sec:evaluation}
We evaluate our CoRAST distributed learning framework for managing complex correlated heterogeneous data distribution in edge computing systems. We use the real-world benchmark weather dataset from \cite{wu2022timesnet}  and distribute the dataset's measurement data across clients. The dataset includes a variety of measurements such as dew temperature (Tdew), potential temperature (Tpot), air pressure (p), and relative humidity (rh), among others. It is characterized by temporal patterns, such as variations in temperature and humidity throughout the day, as well as inter-feature correlations, such as the relationship between the pressure and temperature.   

To demonstrate CoRAST's ability to process correlated heterogeneous data,  each client is assigned a distinct subset of data types without any overlap or data sharing. The specific experimental settings are detailed in Table \ref{tab:setting}, where each set of square brackets denotes the data types assigned to a particular client. There are two  settings: Setting 1 (centralized/distributed) and Setting 2 (centralized/distributed).  Setting 2 provides clients with additional climate information (i.e., Tpot and p) compared to Setting 1. 

In  distributed settings, three clients each gather non-overlapping sets of climate measurements, while in centralized settings, one central client collects all  data. Given that clients collect different types of data with possible dimension differences (Setting 2), traditional FL  struggles due to the local models' diversity, making it hard to aggregate these into an effective global model with potentially different architectures.

 For all experimental setups, the server accesses all the client data to train the FM  for  representation learning. In the centralized cases, the learning process is equivalent to training a lightweight local model with a FM-based learning approach pre-trained on the same data. We explore two additional scenarios: (1) \textbf{No FM}, where FM is not utilized, and learning is purely local, akin to traditional multi-variate time-series analysis.
 (2) \textbf{CoRAST-rho}, where the server uses only the density measure \texttt{rho} -- a variable not directly observed but significantly correlated with client data -- to evalute the framework without direct  data sharing from clients.

 \textbf{Setup: } The dataset is divided into training, validation, and test sets in a 7:1:2 ratio. The server's backbone model employs TS2Vec~\cite{yue2022ts2vec}, consisting of 3 contrasting layers, each containing 64 hidden units. It is pre-trained on the training set using a self-supervised approach with hierarchical contrastive loss and finally generates a 256-dimensional global representation for each time step. 
 
 On the client side,  local data is divided into sequences of 128 and processed through a 3-layer temporal convolutional network (TCN)~\cite{bai2018empirical},  with a kernal size of 3 and 64 hidden units. Meanwhile, a fully connected network is applied to the server-derived representation corresponding to the same time step. These outputs are combined and fed into an affiliated output layer to produce final results. Table \ref{tab:setup} summarizes the number of parameters of the client models, the server model, and the dimension of the representation used in the experiment.

 The server model trains at an initial learning rate of 0.001, and client models at 0.0001, using the Adam optimizer and Cosine Annealing scheduler. Training halts if performance drops on the validation set for three consecutive evaluations (i.e., an early stopping criterion with a patience of 3).


\paragraph{Aligned Objective - H2CO Forecast} We first focus on a common objective across all clients: predicting future H2CO levels, though none directly observe H2CO. This variable is highly correlated with client data.   Training losses alongside test errors, are depicted in Figure \ref{fig:results}(a) and summarized in Table \ref{tab:results1}.  CoRAST enhances learning outcomes across all scenarios. When focusing on a singular global objective, CoRAST significantly improves distributed learning by adding additional environmental insights into learned representations. This approach facilitates a more effective aggregation of client data than traditional forecasting techniques, as evidenced by Table~\ref{tab:results1}'s reductions in forecast error in comparison to the centralized case lacking FM integration. 
 Furthermore, the improvements of CoRAST-rho over no FM show the value of the global representation in conveying critical environmental information, previously inaccessible to client-based learning, thereby improving local model performance. {Setting 2 generally gives lower MSE than Setting 1, indicating the value of including more weather data in the prediction.}

\begin{table}
    \centering
    \caption{H2CO Forecast - Test MSE. CoRAST has the lowest test loss in most settings when all clients have the same local task. CoRAST-rho takes advantage of limited global information to outperform the baseline without a FM.}
    \label{tab:results1}
    \begin{minipage}{\columnwidth}
    \begin{center}
    \begin{tabular}{c c c c}\toprule
        Setting &  No FM & CoRAST-rho & CoRAST \\\midrule
        1-centralized &  0.195  & {0.182} & \textbf{0.171}\\
        1-distributed &  0.391  & {0.303} & \textbf{0.201}\\
        2-centralized &  0.075  & \textbf{0.055} & {0.061 }\\
        2-distributed &  0.151  & {0.146} & \textbf{0.063}\\\bottomrule[\heavyrulewidth]
    \end{tabular}
    \end{center}
    \end{minipage}
\end{table}

\paragraph{Multimodal Tasks - Local Forecast} We explore multimodal local objectives  with clients predicting future values of their specific observed variables, which vary in type. For example, in the 2-distributed setting,  three clients forecast variables [Tdew, Tpot], [rh, p], and [sh], respectively. As a result, distinct local models targeting specific client data need to be learned. Figure \ref{fig:results}(b) and Table \ref{tab:results2} show the average training loss across all variables and test MSE for each variable. The results show that CoRAST improves the training process and  model accuracy in both centralized and distributed cases, which  highlights CoRAST's capability to enhance the distributed learning of interrelated tasks by utilizing correlated environmental data through FMs. Comparing CoRAST and CoRAST-rho reveals the distinct influences that environmental variables exert on each other, emphasizing the complexity of correlations within environmental data.


\begin{table}
    \centering
    \caption{Local Forecast -  Test MSE in Distributed Settings with Distinct Local Tasks for clients. CoRAST has the lowest test loss in most settings. CoRAST-rho takes advantage of limited global information to outperform the baseline without a FM.}
    \label{tab:results2}
    \begin{minipage}{\columnwidth}
    \begin{center}
    \begin{tabular}{c c c c c}\toprule
      Setting & Variable &  No FM & CoRAST-rho & CoRAST  \\\midrule
       \multirow{3}{*}{\shortstack[c]{ 1-distributed}} & Tdew &  0.072  & \textbf{0.056} & {0.075 }\\
        &rh &  0.297  & {0.318} & \textbf{0.276} \\
        &sh &  0.077  & {0.072} & \textbf{0.059 }\\\midrule
     \multirow{5}{*}{\shortstack[c]{ 2-distributed}}  & Tdew &  1.953  & \textbf{1.796} & 1.853\\
        &rh &  0.223  & 0.216 & \textbf{0.211} \\
        &sh &  1.945  & \textbf{1.912} & 1.930\\
        &p &  0.151  &  0.159 & \textbf{0.145} \\
         &Tpot &  2.633  & 2.544 & \textbf{2.438}\\\midrule 

    \end{tabular}
    \end{center}
    \end{minipage}
\end{table}

\section{Conclusion and Future Work}\label{sec:conclusion}
This paper introduces CoRAST,  a general FM-based learning  framework specifically designed for analyzing correlated heterogeneous data in CPS and IoT. CoRAST uses a server-based FM to generate global insights through advanced representation learning and enables client models to enhance task performance. It supports asynchronous updates  for continuous learning without requiring system-level retraining, and is flexible in managing diverse local data and customizing client models. Our evaluation demonstrates CoRAST  improves distributed environmental forecasting on a real-world dataset.

Future work will focus on refining the architecture to better support temporal, spatial, and cross-modal correlations. This includes adapting to diverse training datasets and the discrepancy in data available at training versus inference.  Moreover, the current method of using a uniform global representation fails to meet the diverse needs of individual clients. Introducing client-specific adaptors for tailored representations adds complexity but is essential for personalized learning. These adaptors, reliant on client feedback for performance optimization, introduce a new interaction dynamic.

\bibliographystyle{IEEEtran}
\bibliography{IEEEexample,reference}

\end{document}